%% file: main.tex
\documentclass[10pt,twocolumn,letterpaper]{article}

\usepackage[pagenumbers]{cvpr} 

\input{preamble}

\definecolor{cvprblue}{rgb}{0.21,0.49,0.74}
\usepackage[pagebackref,breaklinks,colorlinks,citecolor=cvprblue]{hyperref}
\usepackage{capt-of,etoolbox}

\newcommand{\introfig}{\protect\centering\vspace{-5mm}
\includegraphics[width=\linewidth]{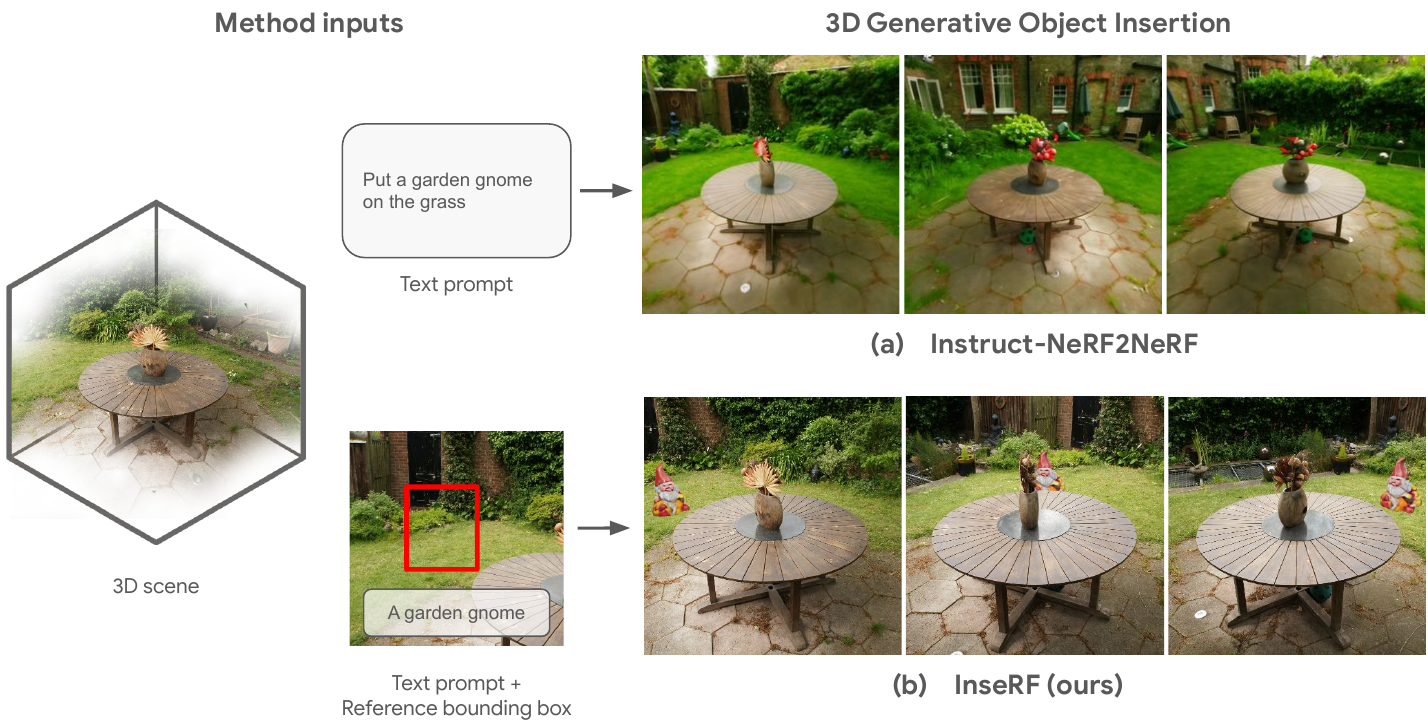}
\vspace{-6mm}
\captionof{figure}{\textbf{Generative insertion of objects in 3D scenes}: 
while Instruct-NeRF2NeRF allows for altering the overall style of scenes or performing edits with a strong positional prior (such as placing a mustache on a face), it fails at inserting objects in arbitrary locations due to the multiview inconsistency of the edits.
Our method, in contrast, is multiview consistent by design and can insert new objects in user-specified locations
}
\label{fig:intro}\vspace{5mm}
}

\def\paperID{***} 
\def\confName{CVPR}
\def\confYear{2024}

\makeatletter
\apptocmd\@maketitle{{\introfig{}\par}}{}{}
\makeatother

\title{InseRF: Text-Driven Generative Object Insertion in Neural 3D Scenes}

\author{
Mohamad Shahbazi$^{1,2*\dagger}$
\qquad Liesbeth Claessens$^{1*}$
\qquad Michael Niemeyer$^{2}$
\qquad Edo Collins$^{2}$ \\
\qquad Alessio Tonioni$^{2}$
\qquad Luc Van Gool$^{1}$
\qquad Federico Tombari$^{2}$ \\
$^{1}$ETH Z\"urich \quad
$^{2}$Google Z\"urich \\
}

\begin{document}

\maketitle

\let\thefootnote\relax\footnote{* Equal contribution.}
\let\thefootnote\relax\footnote{$\dagger$ Work was mainly done during an internship at Google.}

\input{sec/0_abstract}    
\input{sec/1_intro}

\input{sec/2_related_work}
\input{sec/3_method}
\input{sec/4_experiments}
\input{sec/5_conclusion}

{
    \small
    \bibliographystyle{ieeenat_fullname}
    \bibliography{main}
}

\input{sec/X_suppl}

\end{document}

%% file: preamble.tex
\usepackage[dvipsnames]{xcolor}
\usepackage{amsmath}

\DeclareMathOperator*{\argmin}{arg\,min}
\DeclareUnicodeCharacter{200E}{}

%% file: sec/0_abstract.tex
\begin{abstract}
We introduce InseRF, a novel method for generative object insertion in the NeRF reconstructions of 3D scenes. Based on a user-provided textual description and a 2D bounding box in a reference viewpoint, InseRF generates new objects in 3D scenes. Recently, methods for 3D scene editing have been profoundly transformed, owing to the use of strong priors of text-to-image diffusion models in 3D generative modeling. Existing methods are mostly effective in editing 3D scenes via style and appearance changes or removing existing objects. Generating new objects, however, remains a challenge for such methods, which we address in this study. Specifically, we propose grounding the 3D object insertion to a 2D object insertion in a reference view of the scene. The 2D edit is then lifted to 3D using a single-view object reconstruction method. The reconstructed object is then inserted into the scene, guided by the priors of monocular depth estimation methods. We evaluate our method on various 3D scenes and provide an in-depth analysis of the proposed components. Our experiments with generative insertion of objects in several 3D scenes indicate the effectiveness of our method compared to the existing methods. InseRF is capable of controllable and 3D-consistent object insertion without requiring explicit 3D information as input. Please visit our project page at \url{https://mohamad-shahbazi.github.io/inserf}.
\footnote{Correspondence: Mohamad Shahbazi (\href{mailto://mshahbazi@vision.ee.ethz.ch}{mshahbazi@vision.ethz.ch})
.}
\end{abstract}

%% file: sec/1_intro.tex
\section{Introduction}
\label{sec:intro}
Recent advances in the areas of novel view synthesis and generative modeling have led to substantial progress in methods for the generation and manipulation of 3D assets and scenes. Diffusion models~\cite{ho2020denoising, rombach2021highresolution} and their integration with neural reconstruction methods, such as neural radiance fields (NeRFs)~\cite{mildenhall2020nerf}, have enabled the development of powerful 3D generative models for various applications, including text-to-3D ~\cite{poole2022dreamfusion, lin2023magic3d, wang2023prolificdreamer}, single-image-to-3D~\cite{liu2023zero1to3, liu2023one2345, liu2023syncdreamer, qian2023magic123, long2023wonder3d}, 3D shape texturing~\cite{richardson2023texture}, and 3D editing~\cite{instructnerf2023}. 
    
In the particular case of 3D scene editing, recent methods have shown remarkable promise in modifying the style and appearance of real-world scene representations based on textual and spatial guidance. Currently, models that are capable of direct 3D generation and editing are mainly limited to simple and object-centric scenes~\cite{bautista2022gaudi,Chan2021, ntavelis2023autodecoding,liu2023zero1to3}. As a result, for more complex scenes, the majority of the recent editing methods rely on performing edits on different views of the scenes using 2D editing models. One of the most prominent works in 3D scene editing is the recently proposed Instruct-NeRF2NeRF~\cite{instructnerf2023}, an iterative method that performs multi-view edits on the NeRF reconstruction of 3D scenes from textual instructions. Although achieving impressive results, Instruct-NeRF2NeRF is mainly limited to editing the style and appearance of scenes. When prompted to with localized edits or geometry manipulations (such as object removal or insertion) at specified locations, Instruct-NeRF2NeRF often fails to perform the desired edits. This is mainly due to the 3D inconsistency of 2D edits across viewpoints and the lack of proper spatial control.

Recent works have aimed at the 3D-consistent~\cite{dong2023vicanerf, zhuang2023dreameditor} and localized editing~\cite{mirzaei2023watchyoursteps, zhuang2023dreameditor, song2023blending} of 3D scenes. In addition, several studies have specifically tackled object removal and inpainting in 3D scene representations~\cite{Weder2023Removing, spinnerf, mirzaei2023reference, yin2023ornerf, wang2023inpaintnerf360}. However, generating and inserting new objects in scenes in a 3D-consistent way remains an open problem and is mainly limited to cases where edits are strongly constrained by spatial priors (e.g. putting a hat on a head or a mustache on a face). Therefore, in this work, we specifically focus on generative object insertion in 3D scenes, in a way that is consistent across multiple views and placed in arbitrary positions.
    
Generative object insertion in 3D scenes using 2D generative models is a particularly challenging task, as it requires 3D-consistent generation and placement of objects in different viewpoints. A simplistic approach is to separately generate the desired objects using 3D shape generation models~\cite{poole2022dreamfusion, wang2023prolificdreamer} and insert them into the scene using 3D spatial information. However, such an approach requires the accurate location, orientation, and scale of the object in 3D, a non-trivial requirement, especially when in contact with other objects in the scene. Moreover, scene-independent generation of the objects can lead to a mismatch between the style and appearance of the scene and the inserted objects. In this work, we propose a method capable of scene-aware generation and insertion of objects in 3D scenes using the textual description of the objects and a single-view 2D bounding box as spatial guidance.

To circumvent multi-view inconsistencies in appearance and location, the scene-independent generation, and the need for explicit 3D spatial information, we propose grounding the 3D insertion by a 2D view of the object inserted in one reference view of the scene. Given a 3D reconstruction of the scene, we first render a reference view. Then, conditioned on a text prompt and a 2D bounding box, we use an image editing method to add the target object in the reference view. The generated object is then lifted to 3D using a single-view-to-3D object reconstruction method~\cite{liu2023zero1to3, liu2023one2345, liu2023syncdreamer, qian2023magic123, long2023wonder3d}. To place the object in 3D, we propose using the estimated depth of the object in the reference view. After inserting the object in the scene, we perform an optional refinement of the fused scene and objects using the proposed method in Instruct-NeRF2NeRF~\cite{instructnerf2023}.

To evaluate the proposed method, we apply our method to several 3D scenes. Our experiments indicate the ability of the proposed method to insert diverse objects in 3D scenes without the need for explicit 3D spatial guidance. To summarize our contributions:
\begin{itemize}
    \item We address the task of consistent generative object insertion in 3D scenes based on a textual description and a single-view 2D bounding box, which is beyond the capability of the existing 3D scene editing methods
    \item We propose a novel method, based on grounding the insertion using a reference 2D edit, which is capable of 3D-consistent object insertion without requiring explicit information for the 3D placement.
    \item Through our experiments and visualizations, we show the advantage of the proposed method in generative object insertion in comparison to the existing baselines.
\end{itemize}

%% file: sec/2_related_work.tex
\section{Related Works}
\label{sec:related}

\noindent\textbf{Language-based 3D scene editing:} 3D scene editing has recently undergone a considerable transformation by incorporating the strong priors of 2D text-conditioned diffusion models into 3D generative modeling~\cite{instructnerf2023, dong2023vicanerf, zhuang2023dreameditor, song2023blending, mirzaei2023watchyoursteps, park2023ednerf, yu2023editdiffnerf}. Instruct-NeRF2NeRF~\cite{instructnerf2023} proposes an iterative method for 3D scene editing, where different viewpoints of the scene are edited using a text-based 2D editing model and used to fine-tune the scene's NeRF representation. Although highly effective with modifying the existing content, Instruct-NeRF2NeRF often struggles with 3D consistent and localized edits, especially when instructed to remove objects or create new ones in the scene~\cite{dong2023vicanerf, wang2023inpaintnerf360}. To address the view consistency of edits, ViCA-NeRF~\cite{dong2023vicanerf} proposes a method based on a viewpoint-correspondence regularization and a strategy to align the latent space of edited and unedited viewpoints. DreamEditor~\cite{zhuang2023dreameditor} tackles the 3D consistency by adapting the 2D diffusion model to the multi-view images of the scene using DreamBooth~\cite{ruiz2022dreambooth}. DreamEditor additionally identifies a 3D region of interest for localized editing of an existing object based on text-image semantic similarity. The method in~\cite{mirzaei2023watchyoursteps} addresses localized editing differently by obtaining a 3D relevance field for the edits based on the discrepancy between the predictions of the diffusion model with and without instruction conditioning. These methods, despite the improvements, remain limited in their ability to generate new objects, often struggling with cases where a strong spatial prior for the placement of the object does not exist.

\noindent\textbf{Removing objects from 3D scenes:} another direction recently explored in the area of 3D scene editing is 3D-consistent removal and inpainting of objects in the scenes. Some studies assume having multi-view masks of the target object~\cite{Weder2023Removing, mirzaei2023reference}. These multi-view masks, along with other strategies are used to determine the regions to inpaint in different rendered viewpoints. Other studies assume user-provided single-view annotations of the objects and propose approaches to automatically obtain multi-view masks from the reference one~\cite{spinnerf, yin2023ornerf, wang2023inpaintnerf360}. However, such approaches for extracting multi-view masks do not transfer to the task of object insertion, as they rely on the assumption that the objects already exist in the scene.

\noindent\textbf{Generative object insertion:} In contrast to scene stylization and object removal, generating objects in 3D scenes is not well-explored in the existing works. The inpainting method proposed in ~\cite{mirzaei2023reference}, although mainly designed and evaluated for object removal, has been showcased for examples of object insertion as well. To do so, the authors assume multi-view masks of the object are provided, and they propose a method to propagate a single-view inpainting to other viewpoints. However, in addition to requiring multi-view masks as input, the proposed method is mainly limited to forward-facing scenes~\cite{mirzaei2023reference}. FocalDreamer~\cite{li2023focaldreamer} is a concurrent work proposed for adding editable parts to a base 3D shape. Provided with a text prompt and the rough 3D placement of the target edits, FocalDreamer applies score distillation~\cite{poole2022dreamfusion} to add the desired parts to the base shape. Although achieving compelling results, FocalDreamer requires user-provided 3D regions (rotation, translation, and scale), and its generalization beyond base shapes to complex 3D scenes is not investigated. Language-driven Object Fusion~\cite{li2023focalobjecfusion} is another concurrent work that aims at fusing an existing or generated foreground object with a background 3D scene. The authors first adopt a 2D diffusion model for view synthesis from the scene and the object using DreamBooth~\cite{ruiz2022dreambooth}. Then, conditioned on a user-provided 3D bounding box, the authors propose a pose-conditioned dataset update strategy for the training of scene NeRF containing the object. The proposed fusion strategy requires users to provide an exact 3D bounding box. In contrast to the existing language-driven object insertion methods, our approach works well with both forward-facing and 360 scenes, and it only requires a rough 2D bounding box from one rendered view of the scene, making it more suitable for real-world applications.

%% file: sec/3_method.tex
\section{Method}
\label{sec:method}

\begin{figure*}[t]
\centering%
\includegraphics[width=\linewidth]{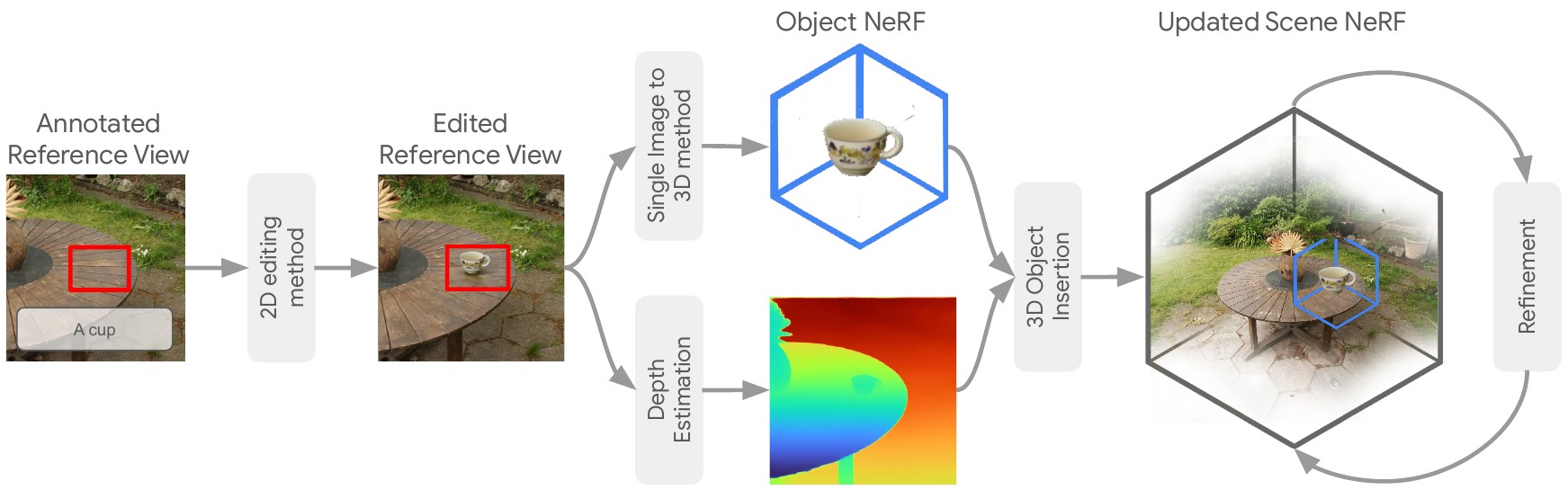}%
\vspace{-2mm}
\caption{\textbf{Overview of the proposed method.} Given a single reference view annotated with a 2D bounding box and a text prompt describing the object to be inserted, a 2D edit is generated portraying a view of the object. This 2D edit is then warped to a 3D model of the object and placed into the scene using the procedure described in section \ref{method:placement}. After the 3D placement, the object and scene representations are fused as described in section \ref{method:fusion}. Finally, an optional refinement can be performed to further improve the appearance.
}
\label{fig_overview}%
\end{figure*}

Our method takes as input a NeRF reconstruction of a 3D scene, a textual description of the target object to be inserted, and a 2D bounding box in a reference rendered view of the scene. As output, our method returns a NeRF reconstruction of the same scene containing the generated target 3D object placed in a location guided by the 2D bounding box. It is noteworthy that our method only requires a rough bounding box, as we rely on the priors of the diffusion models for the exact 2D positioning.

The proposed method consists of five main steps: 1) a 2D view of the target object is created in a chosen reference view of the scene based on a text prompt and a 2D bounding box; 2) a 3D object NeRF is reconstructed from the generated 2D view in the reference image; 3) the 3D placement of the object in the scene is estimated with the help of monocular depth estimation; 4) the object and scene NeRFs are fused into a single scene containing the object in the estimated placement; 5) optionally, a refinement step is applied to the fused 3D representation to improve the insertion further. Fig.~\ref{fig_overview} shows an overview of the proposed pipeline. In the following, we discuss each step in more detail.

\subsection{Preliminaries}
\label{method:pre}

\noindent\textbf{Diffusion Models}
Diffusion models are a type of generative model that maps Gaussian noise to highly realistic and diverse samples. They consist of (1) a forward process that maps data samples $\mathbf{x}_0$ to noise $\mathbf{x}_T$, and (2) a backward process that creates data samples from noise. 

The forward process consists of $T$ steps $t \in [0, T-1]$:
\begin{equation}
q(\mathbf{x}_{t+1}|\mathbf{x}_t) = \mathcal{N}(\mathbf{x}_{t}| \sqrt{1-\beta_{t}}\mathbf{x}_{t-1}, \beta_{t}\mathbf{I}),
\end{equation}
with variances $\beta_t$ chosen such that the noise $\mathbf{x}_T  \sim \mathcal{N}(\mathbf{0}, \mathbf{I})$.

The backward process, which is used to generate data samples from Gaussian noise and optionally an additional conditioning signal, has the following shape:
\begin{equation}
q(\mathbf{x}_{t-1}|\mathbf{x_t}) = \mathcal{N}(\mathbf{x}_{t-1}\mid \mathbf{\mu_{\theta}}(\mathbf{x}_t, t,\mathbf{c}), \mathbf{\sum}_{\theta}(\mathbf{x}_t, t,\mathbf{c})),
\end{equation}
where the parameters of the backward/denoising distributions are predicted by a U-Net, whose weights $\theta$ are optimized by increasing the likelihood of the data samples. Diffusion models can be conditioned on different types of signals, such as images or text, as well as masks, and can be extended for different tasks, such as 2D editing ~\cite{brooks2022instructpix2pix, hertz2022prompt, zhang2023hive} and inpainting~\cite{Lugmayr_2022_repaint, Avrahami_2022_CVPR}.

\noindent\textbf{Neural Radiance Fields}
NeRFs are a novel view synthesis method trained on a set of posed images by minimizing the photometric loss between ground truth and rendered pixels.
A key aspect of NeRF is that pixel colors are not predicted directly. Instead, the density $\mathbf{\sigma}$ and color $\mathbf{c}$ at 3D points in space are predicted by a neural function $f_\phi$. Using these predictions,
the pixel color corresponding to a ray $\vec{\mathbf{r}} = (\vec{\mathbf{o}}, \vec{\mathbf{d}})$ with origin $\vec{\mathbf{o}}$ and viewing direction $\vec{\mathbf{d}}$ can be composed through volumetric rendering. To do so, a set of points along the ray $t_i = \vec{\mathbf{o}} + t \vec{\mathbf{d}} $ is sampled, splitting the ray into a set of intervals $\delta_i = (t_i, t_{i+1}]$. The pixel color of the ray can then be composed as:
\begin{align}
    C(\mathbf{r})  \approx \sum_{i=1}^{N} w_i \mathbf{c_i},\label{eq:nerf_rendering}\\
w_i = T_i(1-\exp{(-\sigma_i \delta_i})), \label{eq:nerf_weights}\\
T_i = \exp{(- \sum_{j=1}^{i-1} \sigma_j \delta_j)} \label{eq:nerf_transmittance}.
\end{align}
In the above equations:
\begin{equation}
    (\sigma_i, \, \mathbf{c}_i) = f\left( \gamma\left(t_i\right); \, \phi\right) , \label{eq:nerf_mlp}
\end{equation}
where the positional encoding function $\gamma$ and the location of the samples $t_i$ depend on the NeRF variant being used.

\subsection{Editing the Reference View}\label{method:2d_edit}
Our editing pipeline starts by choosing one rendered view of the scene as the reference and inserting a 2D view of the target object based on a user-provided text prompt and a 2D bounding box. The reference view is used to ground the 3D insertion by providing a reference appearance and location. Through empirical experiments, we find the additional use of the bounding box important, as the existing text-guided editing methods often struggle with localized 2D object insertions when only receiving spatial guidance from text prompts~\cite{patashnik2023localizing, Zhang2023MagicBrush, zhang2023hive}. To ensure localized 2D insertion within the input bounding box, we opt for a mask-conditioned inpainting method as our 2D generative model. Specifically, we choose Imagen~\cite{saharia2022imagen}, a powerful text-to-image diffusion model, and further adapt it to mask-conditioning by using RePaint~\cite{Lugmayr_2022_repaint}, a method for mask-conditioned inpainting with diffusion models.

\subsection{Single-View Object Reconstruction}
\label{method:123}
After obtaining the reference edit, we extract the 2D view of the object generated within the bounding box and create a 3D reconstruction of it. To do so, we propose exploiting the recent paradigm of single-view object reconstruction using 3D-aware diffusion models~\cite{liu2023zero1to3, liu2023one2345, qian2023magic123, liu2023syncdreamer, long2023wonder3d}. Such reconstruction methods are typically trained on large-scale 3D shape datasets, such as Objaverse~\cite{Deitke_2023_objeverse} and therefore contain strong priors over the geometry and appearance of 3D objects. We use the recently proposed SyncDreamer~\cite{liu2023syncdreamer} for our object reconstruction, as it offers a good trade-off between reconstruction quality and efficiency.

\subsection{3D Placement}\label{method:placement}
\noindent\textbf{Depth Estimation:} The reference 2D bounding box constrains the 3D location of the target object to a frustum in the scene. To determine the location of the object in the 3D frustum, we propose using the prior from monocular depth estimation methods. We apply MiDaS~\cite{Ranftl2022midas} on the edited reference image to estimate the depth of the object with respect to the reference camera. As MiDaS provides non-metric depth measurements, we perform an extra depth alignment between the estimated depth of the edited reference view and the reference depth rendered from the scene NeRF by estimating a global scale and shift between the reference and estimated depth maps. Specifically, to make the alignment more accurate around the object area, we estimate the alignment parameters using weighted least-square estimation, where measurements are inversely weighted based on their distance to the center of the object bounding box (details are provided in the supplementary). After the alignment, we use the depth of the center pixel $d$ in the object bounding box as a rough estimate of the object's center in the frustum, which will be further optimized in the next step.  

\noindent\textbf{Scale and Distance Optimization:} Using the estimated depth $d$ as the distance of the object's center from the reference camera helps with resolving the scale-depth ambiguity of the target 3D object, but it is not accurate enough to closely match the original edit. Additionally, single-view reconstruction methods like SyncDreamer (discussed in Sec. ~\ref{method:123}) are trained to generate multi-view images from fixed camera distance $r'$ and focal length $f'$. In general, as these parameters are different from those of the reference camera, the reconstructed object NeRF appears with a different scale in the reference view once placed at the estimated distance. Therefore, we propose an additional optimization step for the scale and the distance of the object with two constraints: 1) the object must reside at the estimated depth; 2) the rendered view of the object in the reference camera should match the initial edit in scale and appearance. To ensure a proper initial state for the optimization we initialize our scale $s$ and object's distance as:
\begin{align}
    &s_0 = \frac{d}{f'} . \frac{r'}{f} \label{method_s_init} \\ 
    &r_0 = s_0.l + d \label{method_r_init}
\end{align}
where $s_0$ and $r_0$ are the initial object scale and distance, and $l$ is the distance of the 3D point corresponding to the center of the bounding box from the origin of the object NeRF's coordinate system. Given a 3D point $\Vec{P'}$ in the original object NeRF's coordinate system, the corresponding 3D point $\Vec{P}$ in the scaled coordinate system is obtained as:
\begin{equation}
    \Vec{P} = s\Vec{P'}
\end{equation}
To obtain the optimized scale $s^*$ and distance $r^*$, we optimize the Mean Square Error (MSE) between the ground-truth 2D edit $I_G$ and the image $I_R$ rendered using the new parameters:
\begin{equation}
    r^*, s^* = \underset{r,s}{\argmin}~||I_G - I_R||^2
\end{equation}
Fig.~\ref{ablation_optimization} in our ablation study visualizes the effect of scale and distance optimization.

\noindent\textbf{Rotation and Translation:} After obtaining the scale and distance of the object from the reference camera, we proceed to estimate the placement of the object in the scene by estimating its 3D rotation and translation with respect to the camera coordinate system. The origin of the object in the scene's coordinate system is obtained as the point along the ray from the reference camera center passing through the center of the bounding box at the desired distance. To obtain the 3D rotation, we align the x-axis of the object's coordinate system to the vector pointing to the reference camera center from the object's origin.

\subsection{Scene and Object Fusion}\label{method:fusion} Once the location and the orientation of the 3D object in the scene are known, we fuse the NeRF representations of the object and scene to be able to render multi-view images of the scene containing the target object. Given a viewpoint, we transform the rays to the coordinate systems of the scene and the object. Each NeRF representation is applied to the corresponding transformed rays to predict the color and density of the object and scene at each 3D position. To render a viewpoint using the predictions of the two NeRFs, we follow the proposed strategy in ~\cite{Schwarz2020graf}, where the density $\sigma_i$ and color $c_i$ at each 3D point $i$ across a ray in the fused representation are defined as:
\begin{align}
    &\sigma_i = \sigma_i^s + \sigma_i^o \label{eq:sum_density}, \\
    &c_i = \frac{\sigma_i^s c_i^s + \sigma_i^o c_i^o}{\sigma_i^s + \sigma_i^o},\label{eq:color}
\end{align}

where $\sigma_i^s$ and $c_i^s$ are the density and the color of the corresponding sample the scene NeRF, and $\sigma_i^o$ and $c_i^o$ are those of the one in the object NeRF.
To be able to use such formulation in our method for merging the object and the scene, it is crucial to take the scaling of the object's coordinate system into account. Going back to the approximation of the volumetric rendering integration, discussed in Sec. \ref{method:pre}, in equation~\ref{eq:nerf_weights}, $\sigma_i\delta_i$ can be seen as the Riemann approximation of the area under the density curve across the ray at interval $\delta_i$. Simply replacing $\sigma_i$ in equation~\ref{eq:nerf_weights} with the definition in equation~\ref{eq:sum_density} results in an inaccurate estimation of the area under the density curve for the merged representation, as the intervals between every two consecutive samples across the rays are not equal between scene and object coordinate systems due to the scaling of the object coordinate system (discussed in section~\ref{method:placement}):
\begin{equation}
    \delta_i^s = s^* \cdot \delta_i^o, 
\end{equation}
$\delta_i^s$ and $\delta_i^o$ are the intervals in the scene and object NeRFs, respectively, and $s^*$ is the optimized scale obtained in section~\ref{method:placement}. To compensate for the scaling of the intervals, we modify equations~\ref{eq:sum_density} and ~\ref{eq:color} as:
\begin{align}
    &\sigma_i = \sigma_i^s + \frac{\sigma_i^o}{s^*} \\
    &c_i = \frac{\sigma_i^s c_i^s + \sigma_i^o c_i^o/s^*}{\sigma_i^s + \sigma_i^o/s^*},
\end{align}
As we also show in Fig.~\ref{ablation_density} in our ablation study, the proposed modification is necessary for the correct rendering of the fused NeRFs.

\begin{figure*}[t]
\centering%
\includegraphics[width=\linewidth]{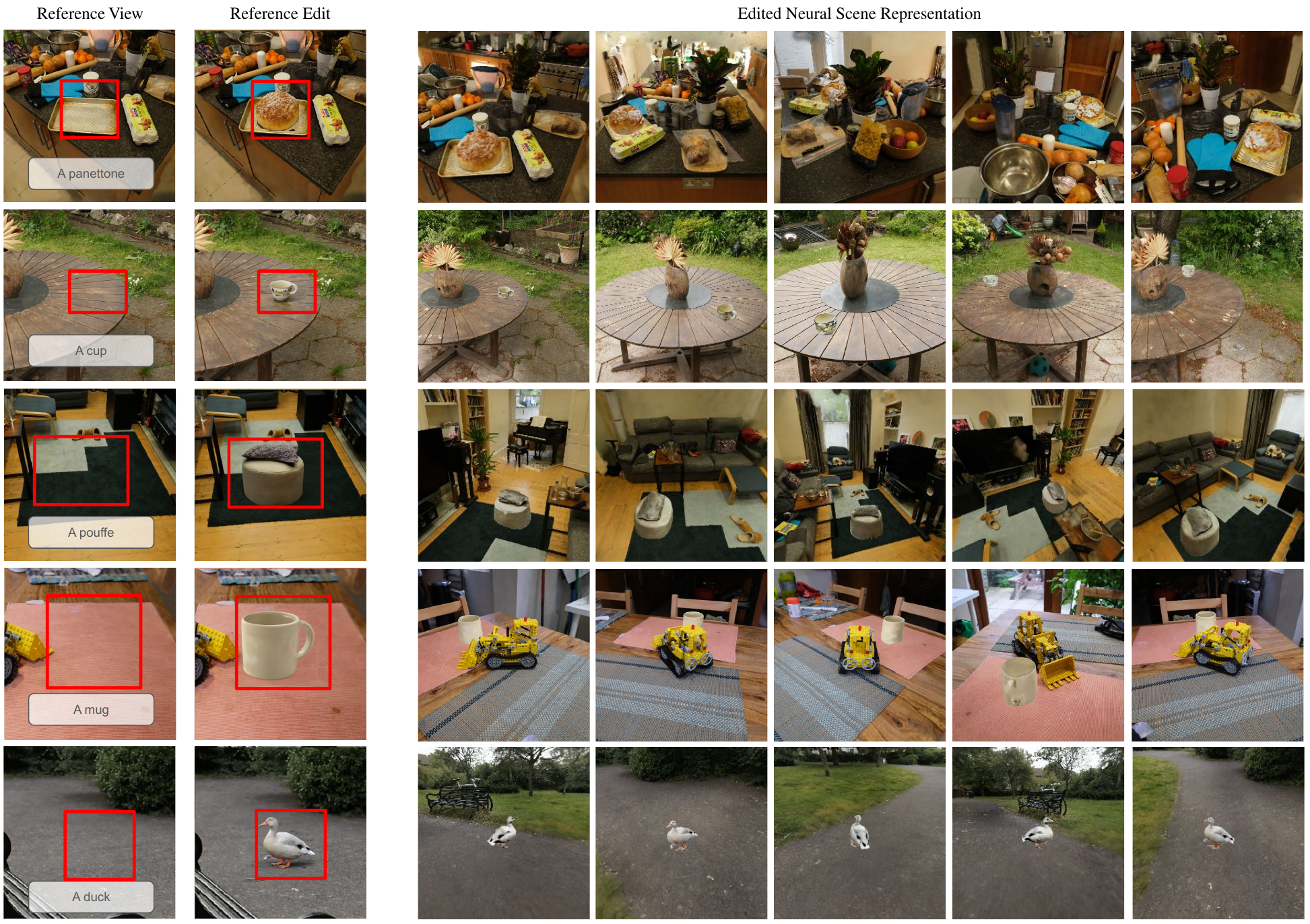}%
\vspace{-2mm}
\caption{Examples of using InseRF to insert an object into the neural representation of different indoor and outdoor scenes.}
\label{fig_visual}%
\end{figure*}

\subsection{Refinement}\label{method:reinement} 
As the final step in our pipeline, we optionally refine the fused scene and object to improve upon the imperfections introduced in the initial reference edit or the single-view reconstruction. To do so, we adapt the iterative refinement proposed in Instruct-NeRF2NeRF~\cite{instructnerf2023} to our setup. First, a set of images is rendered from different viewpoints of the fused NeRF. Then the sampled views are further refined using the 2D diffusion model and added to the optimization of the NeRF consecutively. An important difference between our refinement and Instruct-NeRF2NeRF is that we can obtain multi-view object masks for free from the inserted object to restrict the refinements to the object region. Additionally, in contrast to Instruct-NeRF2NeRF, as the location of the object is known in our refinement step, we adjust our camera trajectory to revolve around the object. We also arrange the sampled viewpoints such that more frontal views are edited and used for NeRF optimization earlier. We find such adjustments to increase the capability of our refinement step. The effect of the proposed refinement is visualized in Fig.~\ref{ablation_refinement} of our ablation study.

%% file: sec/4_experiments.tex
\begin{figure*}[t]
    \centering
    \begin{subfigure}[t]{0.5\textwidth}
        \centering
        \includegraphics[width=\textwidth]{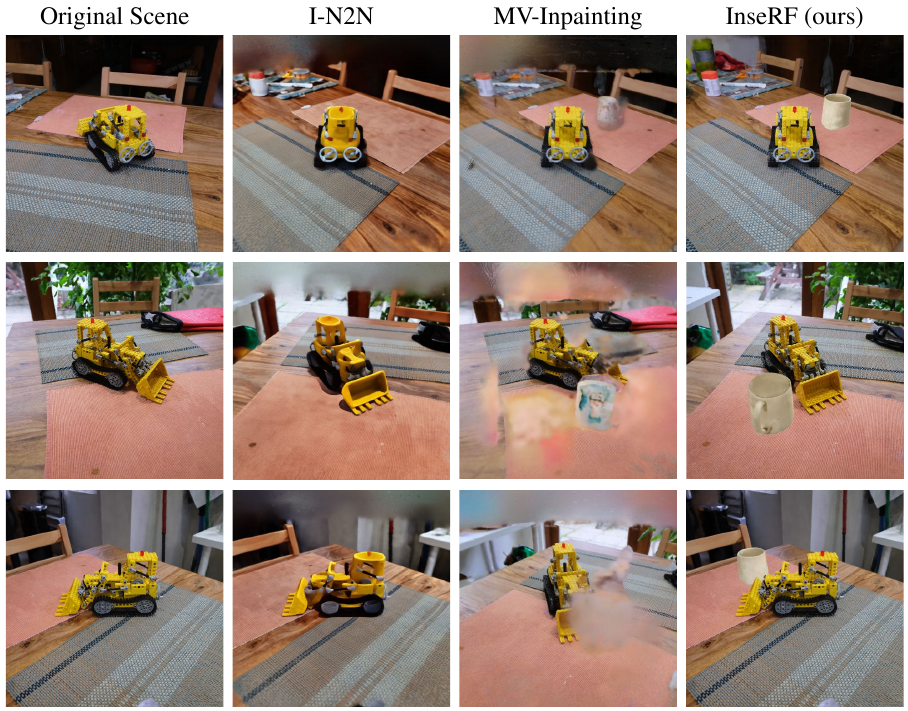}
        \caption{A mug on the table}
        \label{fig_comparison_a}
    \end{subfigure}%
    ~ 
    \begin{subfigure}[t]{0.5\textwidth}
        \centering
        \includegraphics[width=\textwidth]{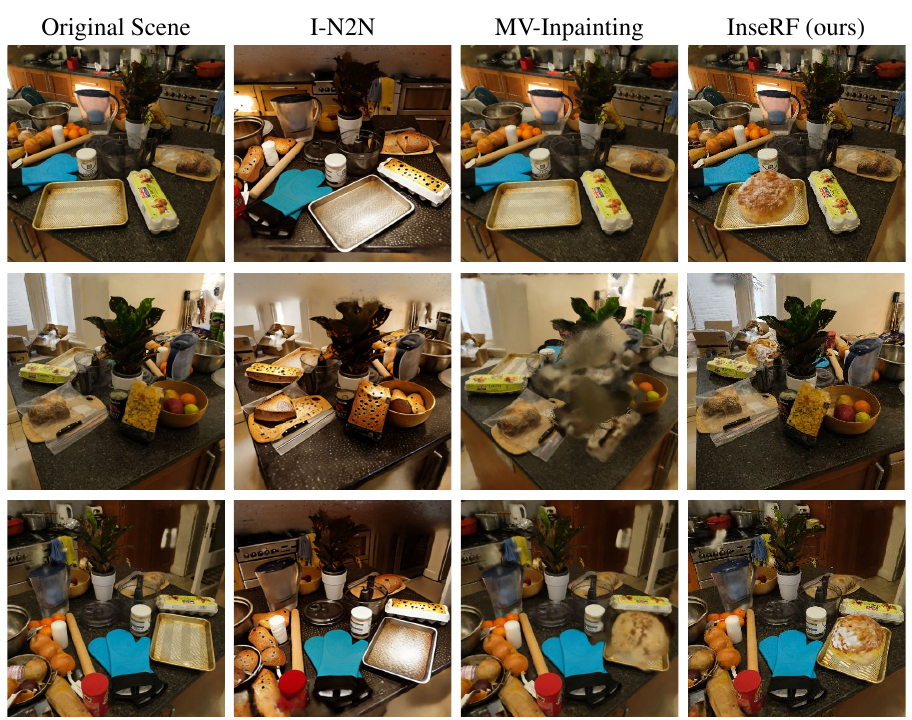}
        \caption{A panettone on the tray}
        \label{fig_comparison_b}
    \end{subfigure}
    \vspace{-2mm}
    \caption{\textbf{Qualitative comparison} of object insertion with different methods. I-N2N modifies existing objects instead of inserting a new object, and the inpainting baseline fails to create geometry at the desired location. Our method, in contrast, can insert new 3D-consistent objects at the desired location.}
    \label{fig_comparison}
\end{figure*}

\section{Experiments}\label{sec:experiments}
In this section, we explain our training and evaluation procedures in more detail. Moreover, we provide the results of our evaluation and comparison with baselines. Finally, we provide an ablation study and analysis of different components of the proposed method.

\subsection{Experimental Details}
\noindent\textbf{Implementation Details:}
For the NeRF representation of objects and scenes, we use MipNeRF-360 \cite{barron2022mip} adapted to the hash grids introduced in Instant-NGP \cite{muller2022instant}. 
For a more exhaustive description of the implementation of our method, we kindly refer the reader to the supplementary materials.

\noindent\textbf{Datasets:}
We evaluate our method on a subset of real indoor and outdoor scenes from datasets proposed in MipNeRF-360 \cite{barron2022mip} and Instruct-NeRF2NeRF~\cite{instructnerf2023}.

\noindent\textbf{Baselines:}\label{exp:baselines}
In our evaluation, we compare the proposed method to the following baselines\footnote{Existing works more related to our method mostly require extra inputs (e.g. 3D boxes) or do not currently provide an implementation~\cite{mirzaei2023reference, li2023focalobjecfusion, li2023focaldreamer}.}:
\begin{itemize}
    \item \textbf{Instruct-NeRF2NeRF (I-N2N)}~\cite{instructnerf2023}: We choose I-N2N as our main baseline, as it is a recent and well-established method for 3D scene editing.
    \item \textbf{Multi-View Inpainting (MV-Inpainting)}: We propose another baseline that follows the refinement strategy in Instruct-NeRF2NeRF, but is additionally provided with accurate multi-view masks for the target object. It is worth emphasizing that, in contrast, our methods only require a rough 2D bounding box in a single reference view.
\end{itemize}
More details on the implementation of our baselines are provided in the supplementary material.

\subsection{Visual Results and Comparisons}
To assess the ability of the proposed method in generative object insertion, we provide visual examples of applying our method to different 3D scenes in Fig.~\ref{fig_visual}. As shown, our method can insert 3D-consistent objects in the scenes. Especially noteworthy is the ability of our method to insert objects on different surfaces, a challenging task in the absence of exact 3D placement information.

In Fig.~\ref{fig_comparison}, we provide a visual comparison with the baselines discussed in Sec.~\ref{exp:baselines}. Attempting to insert new objects in the scene using I-N2N often results in global changes in the scene and modifying existing objects toward the target instead of creating new ones (note how I-N2N changes the Lego truck in~\ref{fig_comparison_a} toward a mug and the items on the kitchen counter~\ref{fig_comparison_b} toward a panettone). Using multi-view masks in the MV-Inpainting baseline helps with limiting the 2D edits to the object region and provides strong spatial guidance. However, 2D edits remain inconsistent from different viewpoints. Therefore, using the edits to optimize the NeRF representation results in 3D floaters and failure to generate the target object in a 3D consistent way. In contrast, our method is capable of localized modification of the scene and of inserting 3D-consistent objects in 3D using only one single-view bounding box as spatial guidance. More visual results are provided in the supplementary material.

\subsection{Ablation and Analysis}

\begin{figure}
\centering%
\includegraphics[width=\linewidth]{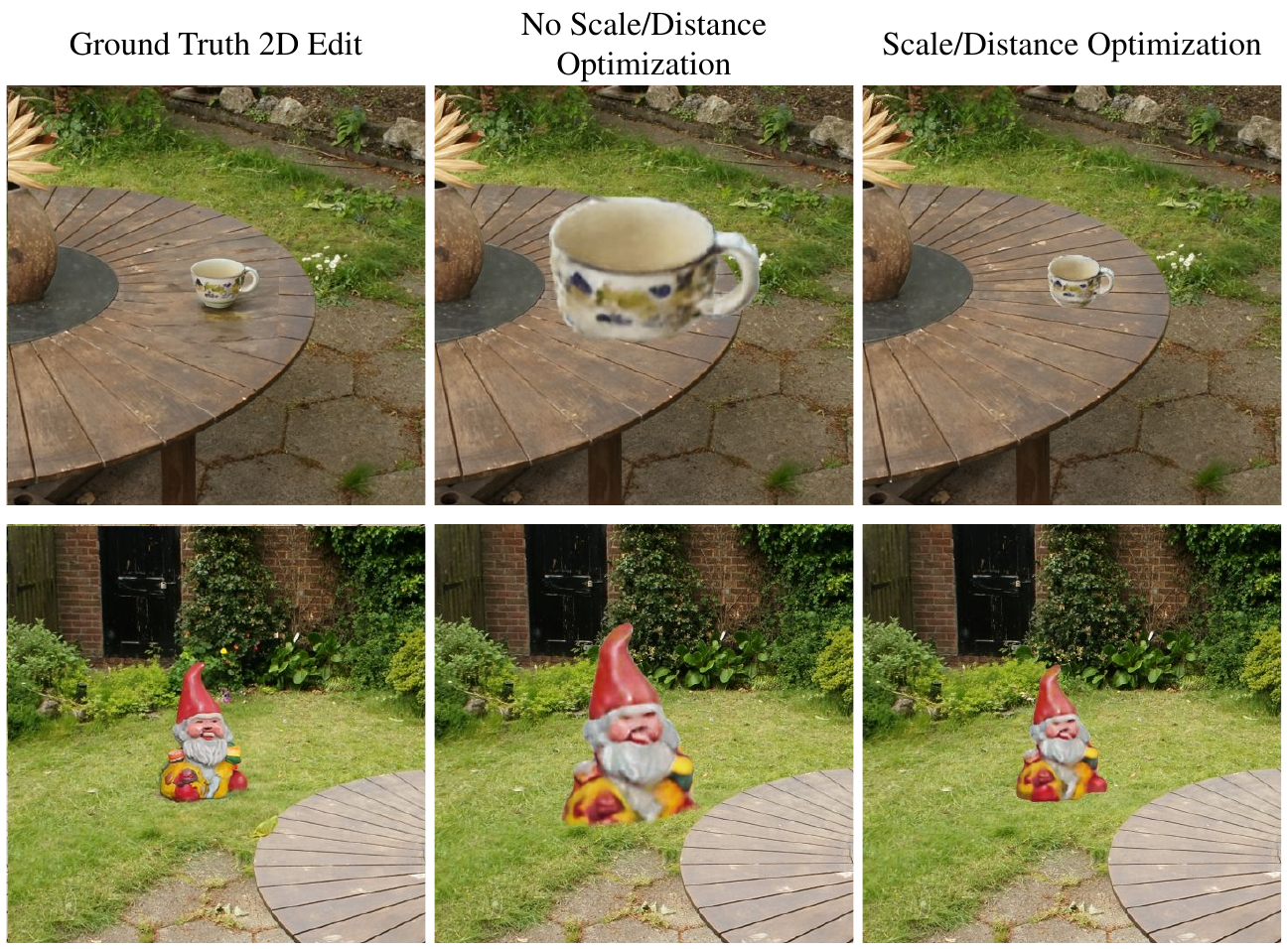}%
\vspace{-2mm}
\caption{Visualization of the effect of \textbf{scale optimisation} on object insertion. The placement of objects is more realistic and faithful to the original edit when performing scale/distance optimization to improve the alignment.}
\label{ablation_optimization}%
\end{figure}

\noindent\textbf{Scale and radius optimization}: In Fig.~\ref{ablation_optimization}, we provide a visual ablation demonstrating the importance of the scale and radius optimization proposed in~\ref{method:placement}, where we compare the placement of the object in the scene using the initial estimation according to Eq.~\ref{method_s_init} and ~\ref{method_r_init} and placement with the extra optimization. As can be seen, the proposed initial estimation would only result in a rough and inaccurate placement of the object. With the proposed optimization, our method can insert objects with the scale and depth matching those of the reference view.

\begin{figure}
\centering%
\includegraphics[width=\linewidth]{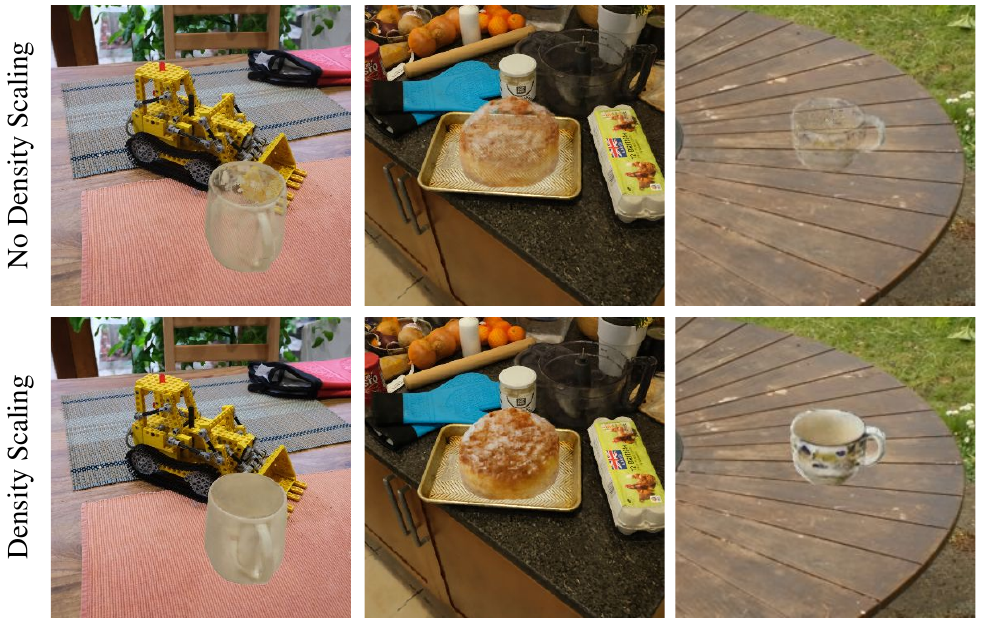}%
\vspace{-2mm}
\caption{Visualization of the effect of \textbf{scaling the densities} when fusing the object and scene representation. When the re-scaling of the object NeRF is not accounted for in the volumetric rendering, the object is not properly displayed in the synthesized views.}
\label{ablation_density}%
\end{figure}

\noindent\textbf{Object density scaling}: In Sec.~\ref{method:fusion}, we proposed an adapted strategy for fusing the NeRF representations of the scene and the object that takes the scaling of the object into account. In Fig.~\ref{ablation_density}, we visualize the importance of our adapted formulation for accurate rendering of the objects inserted in the scene.

\begin{figure}[t]
    \centering
    \begin{subfigure}[t]{\linewidth}
        \centering
        \includegraphics[width=\linewidth]{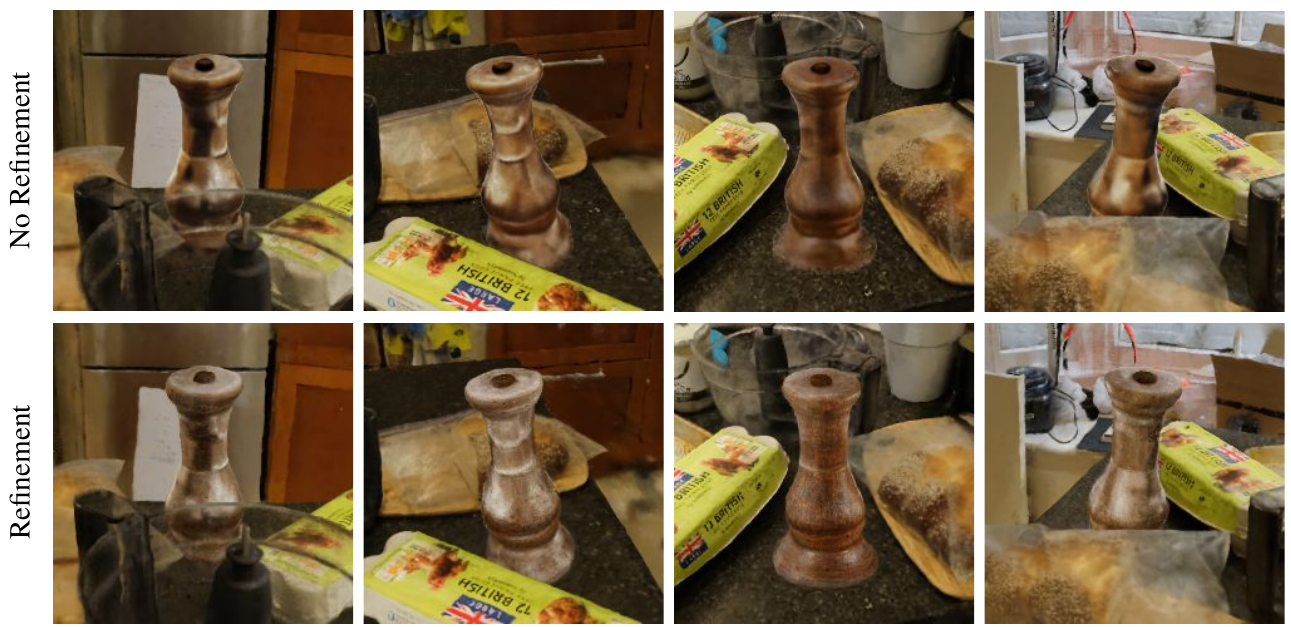}
        \label{fig_refinemnt_peppergrinder}
    \end{subfigure}%
    \vspace{-3mm}
    \begin{subfigure}[t]{\linewidth}
        \centering
        \includegraphics[width=\linewidth]{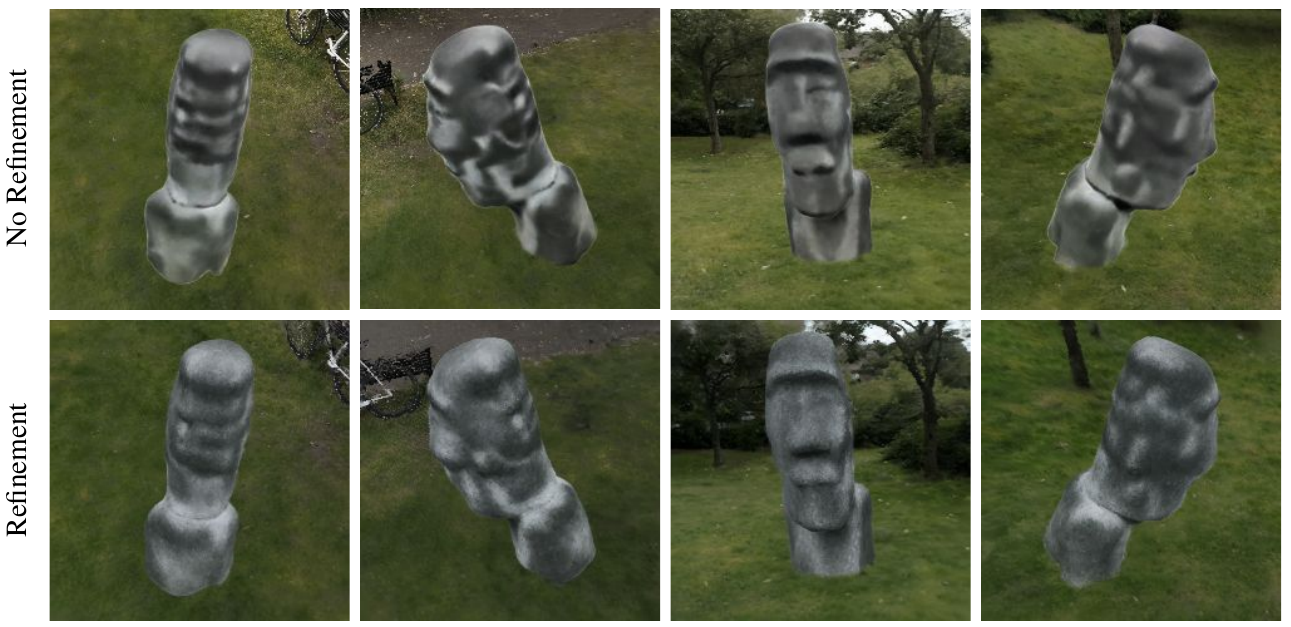}
        \label{fig_refinement_moai}
    \end{subfigure}
    \vspace{-6mm}
    \caption{Visualization of the effect of \textbf{refinement} on object insertion. Our refinement step can add additional texture details and lighting effects.}
\label{ablation_refinement}
\end{figure}

\noindent\textbf{Refinement}: in Sec.~\ref{method:reinement}, we proposed an optional refinement step after inserting the objects in the scenes. Fig~\ref{ablation_refinement} shows examples of the effect of the refinement. As shown, the additional refinement can improve some of the details of the inserted objects, such as the lighting and the texture. 

\subsection{Limitations and Future Work}
Our method is a general pipeline for generative object insertion that is built on top of the existing 2D and 3D generative models and whose parts can be easily swapped. Currently, the performance of our method is limited by the capabilities of the underlying generative models, such as the 2D diffusion model or the single-view object reconstruction method. On the other hand, given our general formulation, future improvements in such models readily transfer to our pipeline.

Our method provides spatial control using a single-view bounding box, as current 2D editing models struggle with the spatial guidance provided in the text prompts. Exploring methods both capable of localized 2D insertion and text-based spatial guidance can lead to improved performance of our whole pipeline. 
Lastly, integrating the concurrently proposed view-consistent editing methods~\cite{li2023focalobjecfusion, dong2023vicanerf} and existing approaches for scene-consistent shadowing and harmonization (e.g. ~\cite{chen2023zeroshot, 10285123}) with our refinement step may bring further improvements to the quality and realism of the insertions.

%% file: sec/5_conclusion.tex
\section{Conclusion}\label{sec:conclusion}
We introduced InseRF, a method specifically designed for generative object insertion in 3D scenes. InseRF takes as input a textual description of the desired object, as well as a 2D bounding box in a single reference viewpoint of the scene. Based on the provided inputs, InseRF generates an object in the 3D scene in a 3D consistent way. To do so, InseRF relies on the priors of 2D diffusion models and single-view object reconstruction methods. The proposed method includes different steps necessary to integrate such methods for the task of in-scene object generation. Through evaluations and visualizations on different 3D scenes, we showed the ability of InseRF in the 3D-consistent generation of objects in the scene without requiring explicit 3D placement information.

%% file: sec/X_suppl.tex
\clearpage

\appendix
\section*{Appendix}
In the appendix, we provide additional visual results, a quantitative evaluation of our method, and an in-depth discussion of the implementation.

\section{Additional Visual Results}

\begin{figure*}[t]
\centering%
\includegraphics[width=\linewidth]{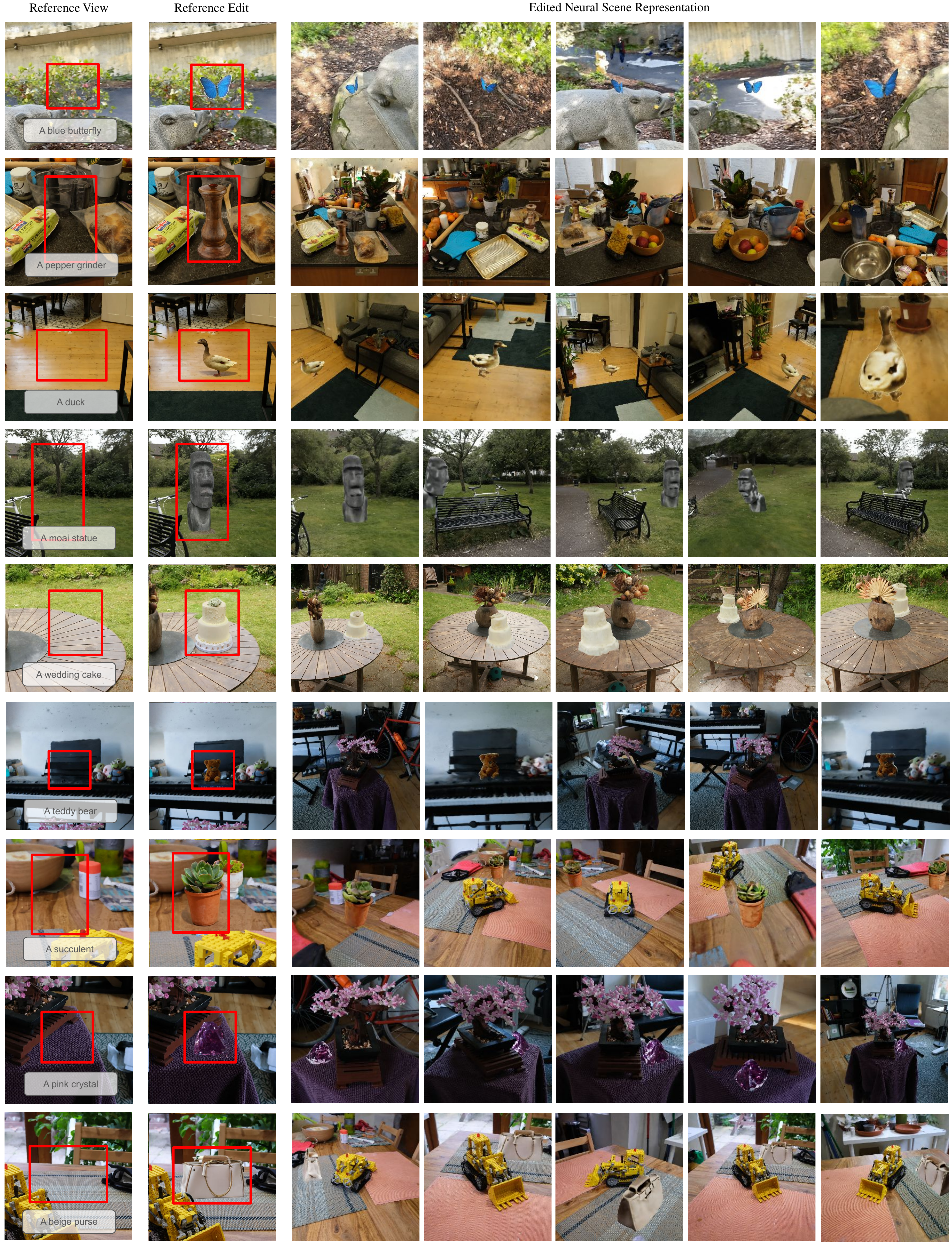}%
\vspace{-2mm}
\caption{
Examples of using InseRF to insert an object into the neural representation of different indoor and outdoor scenes. More examples can be found in Fig.\ \ref{fig_visual} of the main paper.
}
\label{fig_visual_sup}%
\end{figure*}

\noindent\textbf{Visual examples:} in Fig.\ \ref{fig:intro} and \ref{fig_visual} of the main paper, we provided examples of generative object insertion in 3D scenes using our proposed method. Here in Fig.\ \ref{fig_visual_sup}, we provide more visual examples showing the ability of our method to generate objects in 3D scenes.

\begin{figure*}[ht]
\centering
  \begin{subfigure}[t]{0.5\textwidth}
    \centering
    \includegraphics[width=\textwidth]{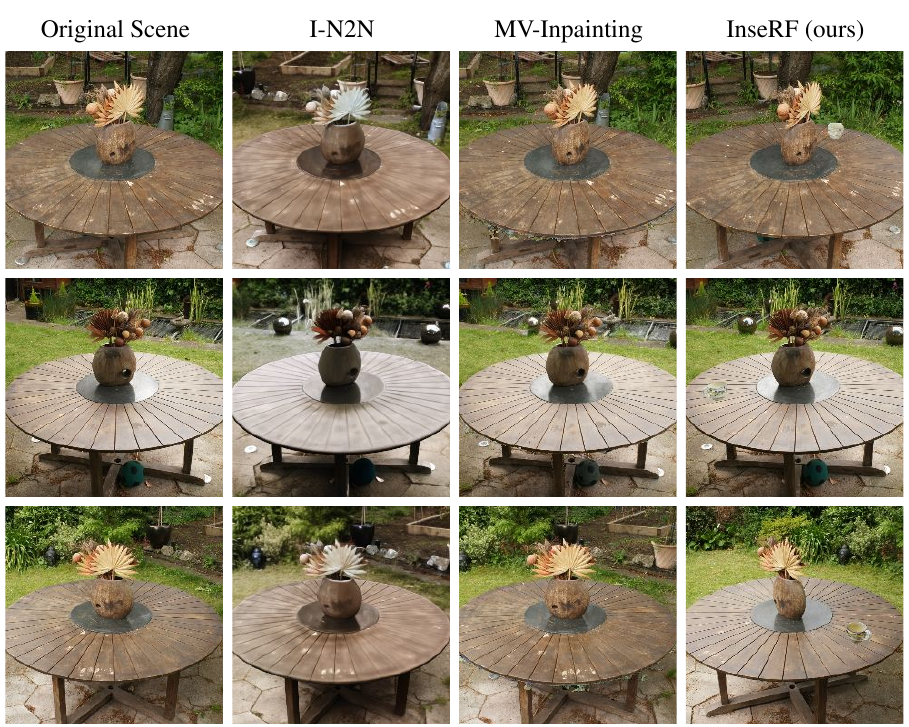} 
    \caption{A cup on the table}
    \vspace{10ex}
  \end{subfigure}
  ~
  \begin{subfigure}[t]{0.5\textwidth}
    \centering
    \includegraphics[width=\textwidth]{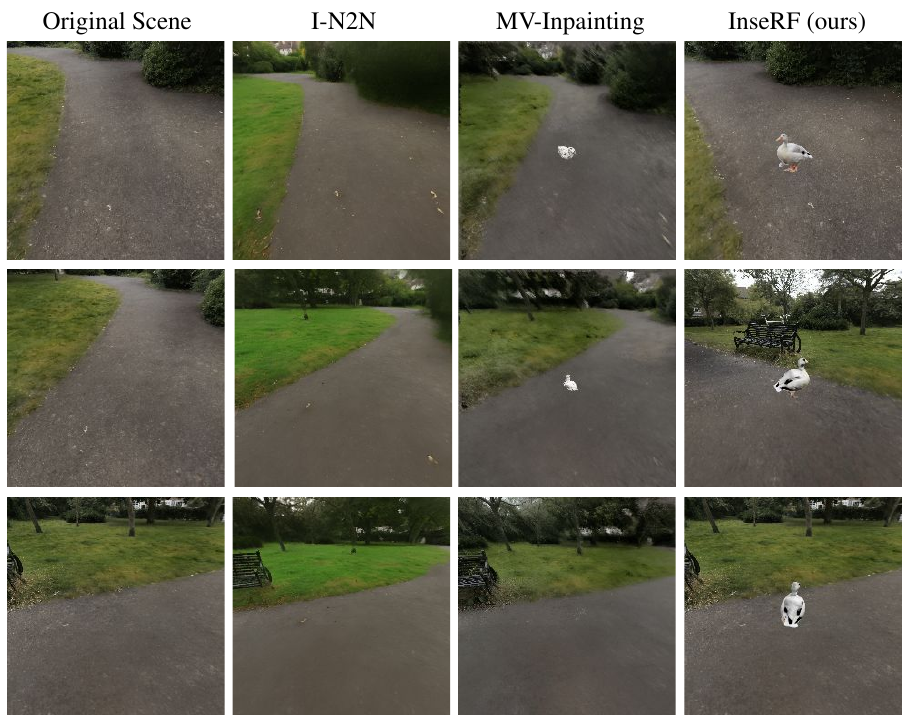} 
    \caption{A duck on the road} 
    \vspace{10ex}
  \end{subfigure} 
  \begin{subfigure}[t]{0.5\textwidth}
    \centering
    \includegraphics[width=\textwidth]{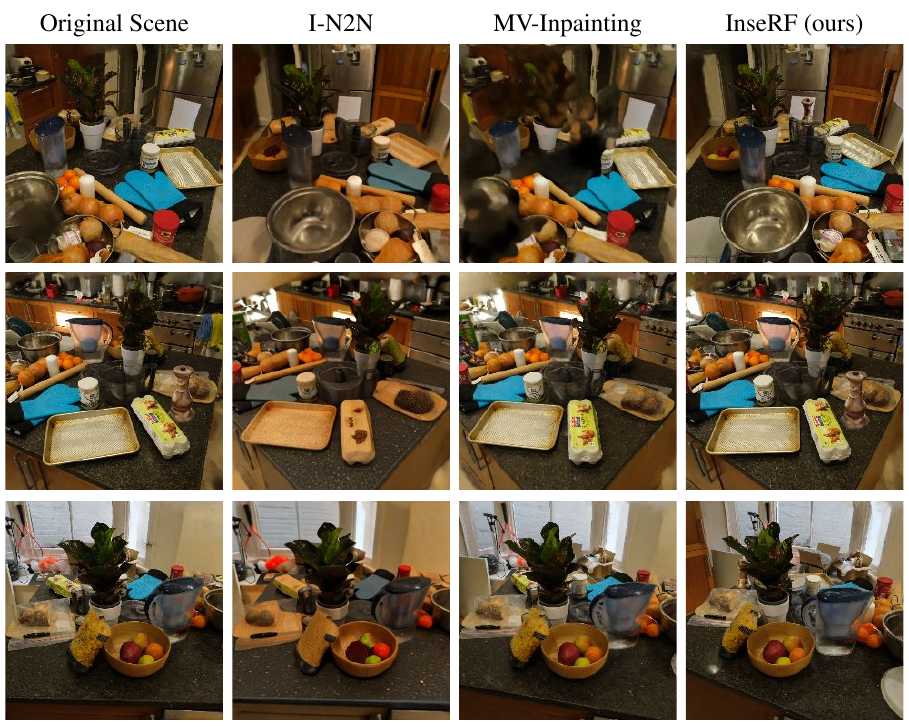} 
    \caption{A pepper grinder on the counter} 
    \vspace{10ex}
  \end{subfigure}
  ~
  \begin{subfigure}[t]{0.5\textwidth}
    \centering
    \includegraphics[width=\textwidth]{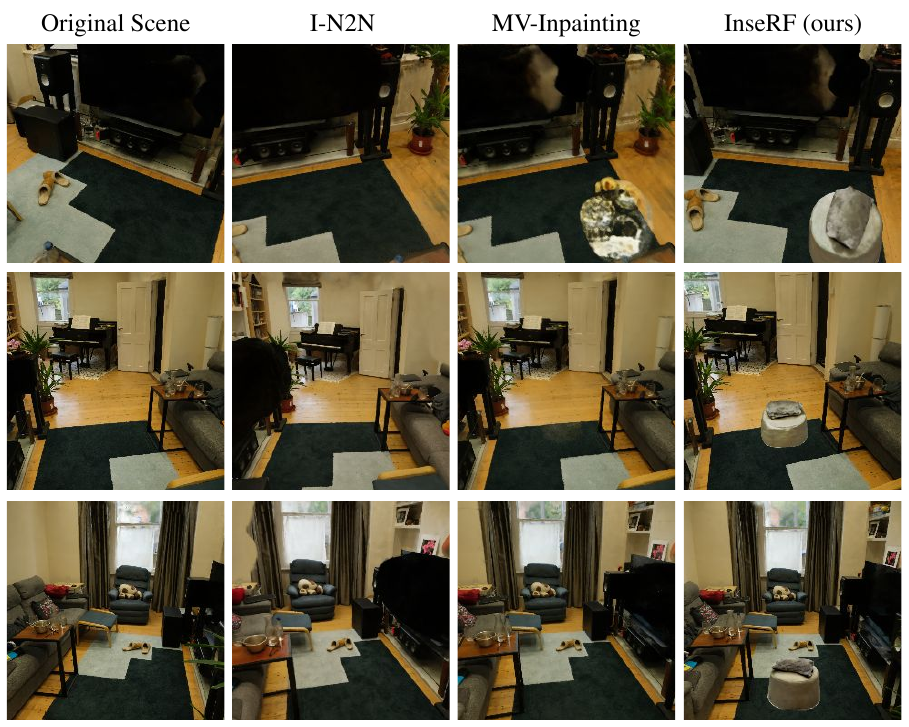} 
    \caption{A pouffe on the carpet} 
  \end{subfigure} 
  \centering
  \caption{Qualitative comparison of object insertion with different methods. I-N2N and multiview inpainting both fail at inserting the geometry of the object at the desired location. Our method, in contrast, can insert new 3D-consistent objects at the desired location. More examples can be found in Fig.\ \ref{fig_comparison} of the main paper.}
  \label{fig_comparison_supp} 
\end{figure*}

\noindent\textbf{Comparison to the baseline}: in Fig.\ \ref{fig_comparison} of the main paper, we provided visual comparisons between the proposed method and our baselines (introduced in Sec.\ \ref{exp:baselines} of the main paper). Fig.\ \ref{fig_comparison_supp} here shows more comparisons with the baselines for a better assessment. As depicted, the two compared baselines struggle with creating the target objects in the scene.

\begin{figure*}
\centering%
\includegraphics[width=\linewidth]{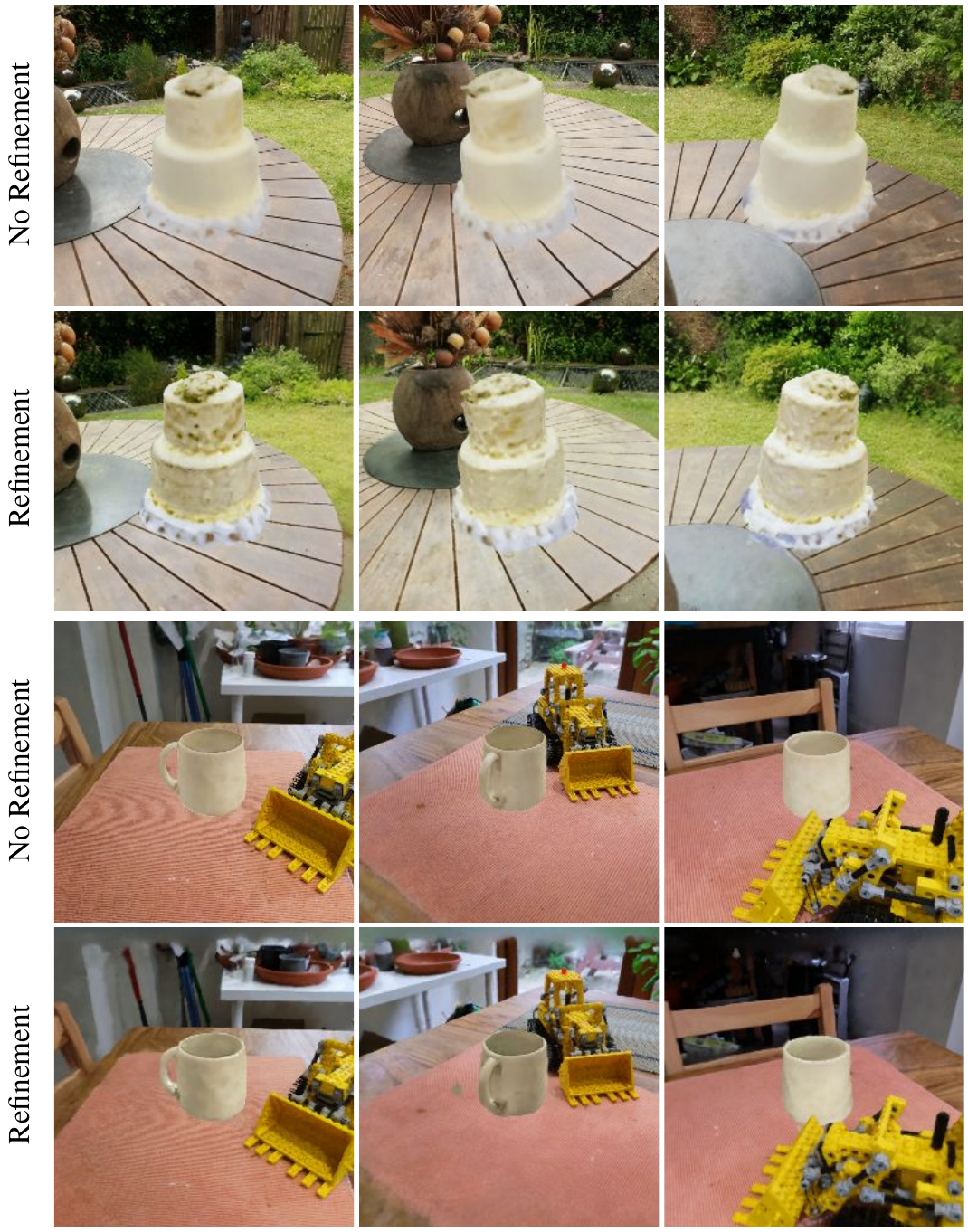}%
\vspace{-2mm}
\caption{\textbf{The refinement step} proposed in our pipeline can improve the texture and the details of the inserted objects, leading to the higher quality and realism of the insertions. More examples can be found in Fig.\ \ref{ablation_refinement} of the main paper.}
\label{fig_refinement_supp}
\end{figure*}

\noindent\textbf{Refinement}: in Fig.\ \ref{ablation_refinement} of the main paper, we provided a visual ablation on the impact of the proposed refinement step in Sec.\ \ref{method:reinement}. Here in Fig.\ \ref{fig_refinement_supp}, we extend the ablation to more examples. As can be seen, the proposed refinement step can improve the texture and details of the inserted objects, resulting in higher-quality and more realistic insertions. For the details of our refinement step, please refer to Sec.\ \ref{method:reinement} of the main paper and Sec.\ \ref{supp_details_refinement} of this supplementary.

\noindent\textbf{Video visualizations:} To better visualize the inserted objects using our method, we additionally provide video visualizations in the supplementary files, showing several examples of our inserted objects, as well as examples of the refinement step.

\section{Quantitative Evaluation}
In addition to the provided visual evaluations, we provide a quantitative evaluation of the proposed method and its comparison with our baselines. Following Instruct-NeRF2NeRF~\cite{instructnerf2023}, we evaluate the methods using three different metrics:
\begin{itemize}
    \item \textbf{CLIP Text-Image Similarity (Text-Image)}: the cosine similarity between the CLIP~\cite{pmlr-v139-radford21a} embeddings of the edit prompt (e.g. "A blue cup") and the images rendered from different viewpoints of the edited scene (We exclude the views where the inserted objects are occluded).
    \item \textbf{Directional Text-Image Similarity (Directional)}: Given a textual description of the original scene (e.g. "A kitchen counter") and an edit prompt describing the scene and the edit (e.g. "A kitchen counter with a mug on top"), this metric measures the similarity of the direction of change from the original scene to the edited one between the image and text CLIP embeddings.
    \item \textbf{Temporal Direction Consistency (Temporal)}: Given two adjacent rendered viewpoints of original and edited scenes, this metric measures how much the change of image embeddings between the two viewpoints in the edited scene is consistent with the one in the original scene.
\end{itemize}

We provide the results of our quantitative evaluation on 8 different edits (5 different scenes) in Tab.\ \ref{tab_quantitative}. All three metrics are based on Cosine similarity, which ranges from -1 to 1. We bring the values between 0 to 1 (the higher the better) for ease of comparison. As depicted, our method effectively outperforms the baselines in the three evaluated metrics. It is worth discussing that, although the provided metrics indicate the advantage of our method over the baselines, we refer the readers to the qualitative results for a better assessment of the evaluated methods. As also highlighted by Instruct-NeRF2NeRF~\cite{instructnerf2023}., the metrics above, although helpful, do not fully capture the effectiveness of methods in 3D scene editing. Exploring alternative metrics that better measure such edits would be an important direction for future studies.

\begin{table}[b]
\caption{Quantitative evaluation of InseRF and its comparison with the baselines on three different metrics proposed in~\cite{instructnerf2023}. For ease of comparison, we report the values (cosine similarities ranging from -1 to 1) after bringing them between 0 and 1. Our proposed method effectively outperforms the baselines in all three metrics.}
\label{tab_quantitative}%
\centering%
\resizebox{\linewidth}{!}{%
\begin{tabular}{llll}
\toprule
\multicolumn{1}{l}{\bf Method} &\multicolumn{1}{c}{\bf Text-Image $\uparrow$}&\multicolumn{1}{c}{\bf Directional $\uparrow$} &\multicolumn{1}{c}{\bf Temporal $\uparrow$} \\ 
\midrule
\multicolumn{1}{l}{I-N2N~\cite{instructnerf2023}} &\multicolumn{1}{c}{0.610} &\multicolumn{1}{c}{0.515} &\multicolumn{1}{c}{0.637} \\
\multicolumn{1}{l}{MV-Inpainting} &\multicolumn{1}{c}{0.606} &\multicolumn{1}{c}{0.499} &\multicolumn{1}{c}{0.724} \\
\multicolumn{1}{l}{InseRF (ours)} &\multicolumn{1}{c}{\bf 0.618} &\multicolumn{1}{c}{\bf 0.545} &\multicolumn{1}{c}{\bf 0.805} \\
\bottomrule
\end{tabular}}
\end{table}

\section{Implementation Details}
\subsection{Inpainting with RePaint}
As mentioned in Sec.\ \ref{method:2d_edit} of the main paper, to generate a 2D view of the target object in the reference view, we condition our diffusion model on a bounding box using RePaint~\cite{Lugmayr_2022_repaint}. Repaint is a training-free inpainting method for pretrained diffusion models that is capable of adding new content to an image in the regions specified by an arbitrary binary mask. Repaint primarily consists of 2 components: 1.) mask conditioning and 2.) re-sampling.

To enable mask conditioning, in every step $t$ of the diffusion process, RePaint applies a mask-based blending to the output $x_{t-1}$ as follows:
\begin{align}
    &x_{t-1} = (1-M) \odot x_{t-1}^{known} + M \odot x_{t-1}^{unknown}
\end{align}

where $x_{t-1}^{known}$ is sampled using known pixels in the given
image, $x_{t-1}^{unknown}$ is sampled from the model given the previous iteration $x_t$, and $M$ is the binary mask. $\odot$ denotes element-wise multiplication. In our setup, we set $M$ to be the area inside the condition bounding box and $x_{t-1}^{known}$ to be noisy versions of the reference image $x_0$ obtained using the forward diffusion process (Eq.\ 1 in the main paper).

When only applying the mask-based blending, the authors of RePaint observe that, although the inpainted region matches the texture of the neighboring region, it is not well-harmonized in the image. Therefore, an additional re-sampling step is proposed, where the blended noisy images go through a few forward diffusion steps and are denoised again, to increase the harmonization of the inpainted regions. The proposed re-sampling step is characterized by two hyperparameters: 1) \emph{jump length}: the number of applied forward diffusion steps; 2) \emph{steps}: the number of repetitions of adding noise and de-noising of the blended images. In our experiments, we set both parameters to the value 2. 
\subsection{Baselines}
\noindent \textbf{Instruct-NeRF2NeRF (I-N2N)}: For our I-N2N baseline, we created a reimplementation in JAX on top of the Mip-NeRF360 code. Our implementation uses the official pretrained checkpoints of Instruct-Pix2Pix~\cite{brooks2022instructpix2pix} and is compatible with LLFF datasets used in our experiments.

\noindent \textbf{Multi-View Inpainting (MV-Inpainting)}: In Sec.\ \ref{exp:baselines} of the main paper, we proposed a baseline called Multi-View Inpainting (MV-Inpainting). MV-Inpainint is designed to insert objects into a 3D scene given accurate multi-view binary masks at the input. To ensure a fair comparison, MV-Inpainting uses the same 2D editing method as ours (Imagen~\cite{saharia2022imagen} with RePaint~\cite{Lugmayr_2022_repaint}) to generate the target object in each viewpoint within the corresponding mask. In contrast to I-N2N, MV-Inpainting is equipped with localized editing to specifically investigate the importance of 3D consistency between different edited viewpoints.

To obtain the multi-view masks required for MV-Inpainting, we first generate and insert an object in the scene using our proposed object insertion. Then, we extract the multi-view masks of the target object by rendering the 3D object into the training viewpoints. The extracted masks are then used as input to MV-inpainting along with the corresponding text prompt. We would like to emphasize that our method only requires a single-view rough bounding box, in contrast to the multi-view accurate masks in MV-Inpainting.

\subsection{Depth Estimation}
As discussed in Sec.\ \ref{method:placement} of the main paper, we use the monocular depth estimated by MiDaS~\cite{Ranftl2022midas} to determine the location of the target object in the 3D frustum formed by the input bounding box in the reference image. As the estimated depth using MiDaS is non-metric, we perform a global affine depth alignment with the reference depth from the scene's NeRF reconstruction, which we explain in greater detail in the following.

Let $D_R$ be the depth of the reference viewpoint rendered from the scene NeRF (not containing the object), and $\hat{D}_E$ be the estimated depth of the edited reference view (containing the 2D object) using MiDaS. We define the aligned depth map $\hat{D}_A$ of the edited reference view as:
\begin{align}
    \hat{D}_A = a \cdot \hat{D}_E + b
\end{align}
where $a$ and $b$ are the scalar parameters of a global affine transformation. $a$ and $b$ are estimated by solving the following weighted least-square estimation:
\begin{align}
    \underset{a, b}{\min}~\underset{i}\sum\underset{j}\sum(1-M^{(i,j)})  \cdot W^{(i,j)} \cdot (D_R^{(i,j)} - \hat{D}_A^{(i,j)})^2,
\end{align}
where $M$ is a binary mask corresponding to the reference bounding box. For a 2D matrix $A$, $A^{(i,j)}$ denotes the element at row $i$ and column $j$. $W$ is the matrix containing pixel-wise weights for the estimation, negatively correlated with the distance of the pixel from the center of the bounding box located at row $i_c$ and column $j_c$:
\begin{align}
&W_{ij}=1-\sqrt{(i-i_c)^2+(j-j_c)^2}/z, \\
&z = \max(\sqrt{(i-i_c)^2+(j-j_c)^2}),\\
&i \in \{0, ..., h-1\}~\&~j \in \{0, ..., w-1\},
\end{align}
where $z$ is a normalization term, and $h$ and $w$ are the height and width of the reference image, respectively. The weighted estimation of the alignment parameters helps with a more accurate alignment in the region surrounding the inserted object. In practice, we perform our alignments on image crops containing the object and its surroundings instead of the full image.

After aligning the estimated depth map, in order to determine the location of the object in the 3D scene, we first roughly estimate the distance of the center of the object from the camera center to be equal to the depth value at the center of the bounding box $d$. Then, we perform the scale and distance optimization proposed in Sec.\ \ref{method:placement} of the main paper, with the constraint that the depth of the center of the object's rendered view from the reference viewpoint must be equal to $d$ (please refer to discussion on the scale and distance optimization in Sec.\ \ref{method:placement} of the main paper for more details).

\subsection{Rotation and Translation}
Here we provide more details on the process of calculating the rotation and translation of the target object in the scene, discussed in Sec.\ \ref{method:placement} of the main paper. Specifically, we obtain the 3D location $\vec{p}_c$ of the center of the object in the 3D scene as the point along the normalized direction $\vec{v}$ pointing from the camera center to the center of the reference bounding box:
\begin{align}
    &\vec{p}_c = \Vec{o} + r^* \cdot \vec{v}
\end{align}
where $r*$ is the optimized distance obtained from the scale and radius optimization (explained in Sec.\ \ref{method:placement} of the main paper). 

We use the right-handed coordinate system convention for our scene and object NeRFs and place the object in an upward position in the scene centered at $\vec{p}_c$. Moreover, we align the reference view of the object in its coordinate system (corresponding to zero azimuth and elevation) with the reference camera viewpoint in the scene's coordinate system. In other words, we define the axes of the object coordinate system in the scene's coordinate system as follows:
\begin{align}
    &\vec{u}_{object} = [0, 0, 1]^T, \\
    &\vec{x}_{object} = -\vec{v}, \\
    &\vec{y}_{object} = normalize(\vec{u}_{object} \times \vec{x_{object}}), \\
    &\vec{z}_{object} = normalize(\vec{x}_{object} \times \vec{y_{object}}),
\end{align}
The rotation $R$ and the translation $\vec{t}$ are then obtained as:
\begin{align}
    &R = [\vec{x}_{object}, \vec{y}_{object}, \vec{z}_{object}]^T \\
    &\vec{t} = -R\vec{p}_c
\end{align}
Using the obtained rotation, translation, and optimized object scale $s^*$, a point $\vec{p}$ in the scene's coordinate system can be mapped to a point $\vec{p'}$ in the object's one as follows:
\begin{align}
    \vec{p'} = \frac{1}{s^*}[R, \vec{t}] \vec{p} 
\end{align}

\subsection{Scene and Object Fusion}
In Sec.\ \ref{method:fusion} of the main paper, we provided a detailed discussion on how the scene and object NeRFs are fused in our method. In practice, object NeRFs fused in the scene may be queried with points in the 3D space that have not been seen during the object NeRF optimization, resulting in unwanted artifacts. To prevent such artifacts, we consider a 3D bounding box around the inserted objects, setting the density of the points sampled outside to zero. The dimensions of the 3D bounding box are determined based on the camera radius used in the single-view object reconstruction step and are fixed across edits and scenes.

\subsection{Refinement}\label{supp_details_refinement}
In Sec.\ \ref{method:reinement} of the main paper, we proposed an optional refinement based on the iterative NeRF optimization proposed in Instruct-NeRF2NeRF with two modifications: 1) using the multi-view masks obtained from the inserted object to make the refinement localized and 2) sampling viewpoints on a sphere encapsulating the inserted object in the scene. In particular, we sample the viewpoints on a sphere with the radius $r^*$ (the optimized object distance) from the object's center $\vec{p}_c$. Such a sampling strategy allows for better edits by the 2D diffusion model. Moreover, instead of randomly picking the next viewpoint to edit and include in the NeRF optimization, as done in Instruct-NeRF2NeRF, we order the viewpoints in a way that more frontal views are selected first. For example, viewpoints $(azimuth, elevation)$ sampled from $n$ equally-distanced azimuths with step size $\Delta_{theta}$ and $m$ equally-distanced elevations with the step size $\Delta_{phi}$ are arranged as an ordered set $V$:
\begin{align}
    &V=\{(i \cdot \Delta_{\theta}, j \cdot \Delta_{\phi}))~|~i \in I~\&~j \in J\} \\
    &I = \{0, 1, -1..., n/2, -n/2\}, \\
    &J = \{0, 1, -1, ..., m/2, -m/2\},
\end{align}

Such ordering improves the 3D consistency of the refinement step, as it decreases the conflict caused by randomly selected and independently edited viewpoints.